\pdfoutput=1
\documentclass[letterpaper, 10pt, conference]{ieeeconf}
\IEEEoverridecommandlockouts
\usepackage{cite}
\usepackage{amsmath,amssymb,amsfonts}
\usepackage{graphicx}
\usepackage[table]{xcolor}
\usepackage{booktabs}
\usepackage{multirow}
\usepackage{microtype}
\usepackage{xspace}
\usepackage{url}
\usepackage{textcomp}
\let\labelindent\relax
\usepackage{enumitem}
\setlist[itemize]{leftmargin=*,nosep,topsep=1pt}
\definecolor{linkblue}{rgb}{0.0,0.2,0.6}
\usepackage[colorlinks,allcolors=linkblue]{hyperref}

\newcommand{\eg}{e.g.,\xspace}
\newcommand{\ie}{i.e.,\xspace}
\newcommand{\etal}{et~al.\xspace}
\newcommand{\ours}{I2CD\xspace}
\newcommand{\SRmjOursC}{99.2}
\newcommand{\SRmjOursL}{0.05}
\newcommand{\SRmjOursP}{0.22}
\newcommand{\SRmjOursR}{99.2}
\newcommand{\SRmjRawC}{40.2}
\newcommand{\SRmjRawL}{2.77}
\newcommand{\SRmjRawP}{0.26}
\newcommand{\SRmjRawR}{83.5}
\newcommand{\SRpbOursC}{99.2}
\newcommand{\SRpbOursL}{0.03}
\newcommand{\SRpbOursP}{0.01}
\newcommand{\SRpbOursR}{100.0}
\newcommand{\SRpbRawC}{40.2}
\newcommand{\SRpbRawL}{2.48}
\newcommand{\SRpbRawP}{0.01}
\newcommand{\SRpbRawR}{78.7}

\newcommand{\SRpbConL}{0.01}
\newcommand{\SRpbConP}{0.01}
\newcommand{\SRpbConR}{100.0}
\newcommand{\SRgeOursC}{99.2}
\newcommand{\SRgeOursL}{3.34}
\newcommand{\SRgeOursP}{0.05}
\newcommand{\SRgeOursR}{99.2}
\newcommand{\SRgeRawC}{40.2}
\newcommand{\SRgeRawL}{9.24}
\newcommand{\SRgeRawP}{0.06}
\newcommand{\SRgeRawR}{100.0}

\newcommand{\SRisOursC}{98.4}
\newcommand{\SRisOursL}{0.18}
\newcommand{\SRisOursP}{0.00}
\newcommand{\SRisOursR}{100.0}
\newcommand{\SRisRawC}{40.5}
\newcommand{\SRisRawL}{0.39}
\newcommand{\SRisRawP}{0.00}
\newcommand{\SRisRawR}{98.4}
\newcommand{\SRisCdC}{26.8}
\newcommand{\SRisCdL}{1.67}
\newcommand{\SRisCdP}{0.00}
\newcommand{\SRisCdR}{99.2}

\newcommand{\SRisSdfL}{3.75}
\newcommand{\SRisSdfP}{0.00}
\newcommand{\SRisSdfR}{86.0}
\newcommand{\SRrawWrong}{60}
\newcommand{\SRgeDecomp}{68}
\newcommand{\SRoursMinC}{98.4}
\newcommand{\SRoursMinR}{99.2}

\title{\LARGE \bf I2CD: Direct Image-to-Convex Decomposition for\\Simulation-Ready Collision Geometry}
\author{\authorblockN{Qian Wang, Liam Merz Hoffmeister, Brian Scassellati, Daniel Rakita}
\authorblockA{Department of Computer Science, 
Yale University}
\authorblockA{\{peter.wang.qw262, liam.merzhoffmeister, brian.scassellati, daniel.rakita\}@yale.edu}
\thanks{This work was supported by Office of Naval Research award N00014-24-1-2124}
}

\begin{document}

\maketitle
\thispagestyle{empty}
\pagestyle{empty}

\begin{abstract}
Physics simulators and motion planners require convex collision geometry, yet image-to-3D generative models output dense, frequently non-manifold visual meshes. Bridging the two today takes a slow, brittle \emph{reconstruct-then-decompose} pipeline of repair, decimation, and approximate convex decomposition. We present \ours, which predicts a convex decomposition directly from a single RGB image. Rather than train a new image-to-3D model, \ours freezes the pretrained Hunyuan3D-2 image-conditioned diffusion transformer and shape decoder and trains only a lightweight cross-attention head (38\,M parameters, under ten GPU-hours) whose learned ``convex-slot'' tokens emit the halfplane parameters of $K$ convex polytopes. The output is compact, convex by construction, and loads into physics engines without any post-processing, in ${\sim}0.5$\,s per image. On $227$ held-out OmniObject3D and Google Scanned Objects instances, \ours attains the highest volumetric IoU among eight reconstruct-then-decompose pipelines while running $6$--$37\times$ faster end-to-end. In a cross-simulator study in MuJoCo, PyBullet, Genesis, and Isaac Sim, every engine uses \ours geometry as delivered, whereas raw generated meshes ``load'' everywhere but are silently replaced by a different collision shape in most cases or need seconds to minutes of per-object preprocessing. On a physical xArm7, \ours produces planner-ready geometry for a $20$-object cluttered scene in $11$\,s versus $328$\,s for the strongest baseline, at comparable pick-and-place execution success ($85$ vs.\ $90$ of $100$ trials).
\end{abstract}

\section{Introduction}
\label{sec:introduction}

Robots plan and simulate on convex collision geometry. Sampling-based motion planners~\cite{sucan2012ompl,lavalle2006planning}, rigid-body simulators~\cite{coumans2016pybullet,todorov2012mujoco,mittal2025isaaclab,Genesis}, and contact-rich manipulation pipelines all rely on fast, robust proximity and penetration queries between convex bodies, computed with algorithms such as GJK and EPA~\cite{gilbert1988fast,van2001proximity}. Objects are therefore represented as \emph{convex decompositions}, \ie unions of convex parts that approximate a triangle mesh, and production simulators routinely build them with tools such as V-HACD~\cite{mamou2009simple} and CoACD~\cite{wei2022approximate}.

Image-to-3D generative models now offer a direct route from a camera image to 3D geometry. The strongest methods sample a high-quality shape latent with a large image-conditioned diffusion or flow-matching model in a single feed-forward pass~\cite{zhao2025hunyuan3d,xiang2025structured,chen2025sam}, and robotics increasingly uses them to build scene geometry for manipulation and planning~\cite{wang2025subsecond,huang2026simulation,agarwal2025scenecomplete}. However, \emph{these models are optimized for visual fidelity, not physical simulation}. They output dense triangle meshes (${\sim}10^6$ faces) with small but near-ubiquitous topological defects such as non-manifold edges and stray fragments (quantified in \S\ref{sec:exp_geometric}). Turning such a mesh into collision geometry takes a fragile chain of decimation~\cite{garland1997surface}, repair~\cite{attene2010lightweight}, watertightening, and approximate convex decomposition. This \emph{reconstruct-then-decompose} workflow~\cite{huang2026simulation} is slow and sensitive to errors at every stage. Skipping it is worse, because it does not fail loudly: as we show in \S\ref{sec:exp_simready}, physics engines accept the raw mesh and silently substitute a single convex hull, producing a plausible-looking scene with the wrong collision shape.

\begin{figure}[t]
  \centering
  \includegraphics[width=0.95\linewidth]{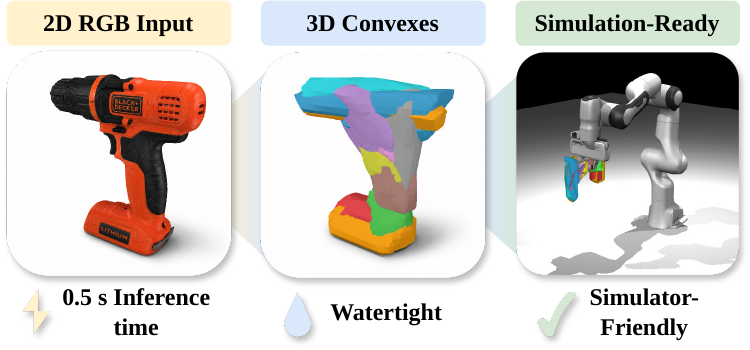}
  \vspace{-4pt}
  \caption{From a single RGB image, \ours predicts a union of convex polytopes in ${\sim}0.5$\,s that physics simulators consume directly, with no mesh extraction, repair, or post-hoc decomposition.}
  \label{fig:teaser}
  \vspace{-3pt}
\end{figure}

We propose \textbf{\ours} (\textbf{I}mage-to-\textbf{C}onvex-\textbf{D}ecomposition), which bypasses this workflow by predicting the convex decomposition directly from a single RGB image (Fig.~\ref{fig:teaser}). Our key insight is that the generator's mesh is the wrong output but its features are the right input: a pretrained image-to-3D model already encodes the object's shape, so only the readout needs to change. We therefore keep the image-conditioned diffusion transformer (DiT) and ShapeVAE decoder of Hunyuan3D-2~\cite{zhao2025hunyuan3d} frozen and replace the mesh readout with a lightweight cross-attention head. Its $K$ learned ``convex-slot'' tokens query the pretrained shape features, and each emits the $H$ halfplanes of one convex polytope in the CvxNet formulation~\cite{deng2020cvxnet}; the union of the $K$ polytopes is the predicted shape. Only this head is trained, with a class-balanced occupancy loss and a convex-pair contrastive loss that suppresses the slab-stacking failure mode of CvxNet-style decoders.

\ours introduces neither a new convex representation, which we inherit from CvxNet, nor a new 3D generator. Our contribution is to show, and validate for robotics, that a modern image-to-3D foundation model becomes a direct source of simulation geometry with a 38\,M-parameter head and under ten GPU-hours of training, inheriting an open-world shape prior that, to our knowledge, no image-to-convex model trained from scratch currently offers. The single-view modes of CvxNet~\cite{deng2020cvxnet} and BSP-Net~\cite{chen2020bsp} are category-specific and have no open-world weights (\S\ref{sec:related}). In summary, we:
\begin{itemize}
  \item adapt a frozen Hunyuan3D-2 backbone to emit $K$ halfplane-parameterized convexes directly, with a training objective that yields compact, spatially coherent parts;
  \item benchmark against eight reconstruct-then-decompose pipelines (Hunyuan3D-2 and SAM3D~\cite{chen2025sam} paired with CoACD, V-HACD, CvxNet, and BSP-Net) on $227$ held-out OmniObject3D~\cite{wu2023omniobject3d} and GSO~\cite{downs2022googlescannedobjectshighquality} objects, where \ours achieves the best volumetric IoU at $6$--$37\times$ lower end-to-end latency;
  \item test \emph{simulation-readiness} beyond ``it loads'' with a drop-and-settle study in MuJoCo, PyBullet, Genesis, and Isaac Sim (PhysX): raw generated meshes are silently replaced by a different shape or require costly preprocessing, while \ours geometry is used as delivered in every engine;
  \item execute $100$ pick-and-place trials on a physical xArm7 in a $20$-object cluttered scene, where \ours cuts the time to planner-ready scene geometry from $328$\,s to $11$\,s at comparable execution success ($85$ vs.\ $90$ of $100$).
\end{itemize}
Code, models, and the benchmark will be released.

\section{Related Work}
\label{sec:related}

\textbf{Image-to-3D generation.}
Single-image reconstruction has progressed from template deformation~\cite{wang2018pixel2mesh} and implicit fields~\cite{mescheder2019occupancy,park2019deepsdf} to large feed-forward generators~\cite{hong2023lrm,xu2024instantmesh} built on diffusion~\cite{ho2020denoising} and flow matching~\cite{lipman2022flow}. The state of the art samples a compact shape latent~\cite{zhang20233dshape2vecset,zhang2024clay} with an image-conditioned DiT and decodes it to a dense mesh~\cite{zhao2025hunyuan3d,xiang2025structured,chen2025sam}. Robotics has begun to use these models to build scene geometry from images~\cite{wang2025subsecond,agarwal2025scenecomplete,huang2026simulation}, but always through their mesh output followed by post-processing. We exploit an asymmetry in these systems: their meshes are unsuited to collision geometry, yet their pretrained shape decoders encode a rich, general-purpose geometric prior that a small head can read out into a different output modality.

\textbf{Approximate convex decomposition (ACD).}
The dominant recipe for collision geometry is to clean a mesh and partition it with V-HACD~\cite{mamou2009simple} or CoACD~\cite{wei2022approximate}; recent variants improve the speed--quality trade-off~\cite{vu2025empart,luo2025rl,knodt2026convex,mastromarino2025voxel}. ACD works well on watertight, well-conditioned meshes. It degrades sharply on the noisy, non-manifold meshes emitted by image-to-3D models, and it takes seconds to tens of seconds per object even on clean inputs (Table~\ref{tab:main_results}).

\textbf{Learned convex decomposition.}
CvxNet~\cite{deng2020cvxnet} represents each part as an intersection of $H$ halfplanes and trains an encoder--decoder on occupancy. BSP-Net~\cite{chen2020bsp} learns binary space partitions, and Neural Parts~\cite{paschalidou2021neural} generalizes to non-convex primitives. Vavilala~\etal~\cite{vavilala2023convex,vavilalaimproved} fit convexes to RGB-D scenes with test-time optimization. Yang~\etal~\cite{learningconvexdecomp2025} supervise a per-point feature field with a contrastive loss derived from straight-line containment in convexes. CvxNet and BSP-Net also offer single-view RGB modes, but these are trained end-to-end per category (ShapeNet) and have no open-world weights, so they cannot handle arbitrary household objects without retraining a full image-to-3D model.

\textbf{Positioning.}
Closest in goal is SPARCS~\cite{huang2026simulation}, which estimates simulation-ready cluttered scenes by jointly optimizing the shapes and poses of a few objects (up to $5$, represented by up to $22$ convex hulls in total) under a shape-differentiable contact model. SPARCS is a scene-level, optimization-based refinement that starts from learned per-object initializations and inherits their convex hulls. \ours is an object-level, feed-forward source of exactly such convex geometry (in $0.5$\,s, for $20$-object scenes). The two are complementary: our output could initialize their optimization, while their physics-aware refinement could correct the pose-registration errors that dominate our robot failures (\S\ref{sec:exp_planning}). \ours is, to our knowledge, the first method to adapt a pretrained image-conditioned 3D \emph{foundation model} to directly output simulation-ready convex geometry. We keep the CvxNet halfplane representation and Yang~\etal's containment principle, but apply the contrastive loss directly to the per-convex soft responsibilities of our head, without an auxiliary feature field.

\section{Method}
\label{sec:method}

\begin{figure}[t]
  \centering
  \includegraphics[width=\linewidth]{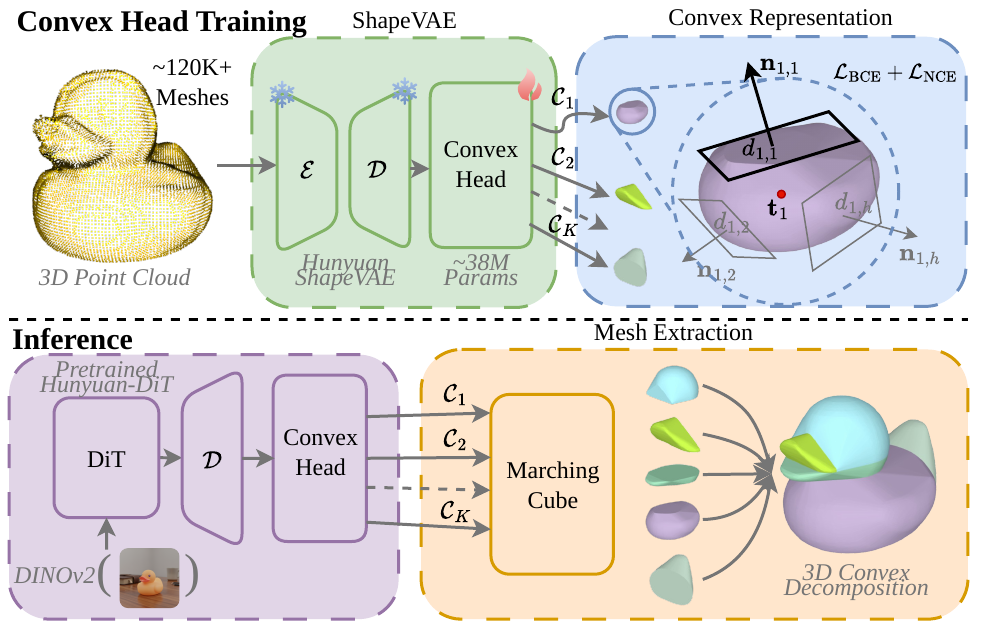}
  \vspace{-14pt}
  \caption{\textbf{Pipeline.} The frozen Hunyuan3D-2 DiT samples a shape latent from the image and the frozen ShapeVAE decoder $\mathcal{D}$ expands it into per-token shape features. Our trainable convex head uses $K$ learned convex-slot tokens that cross-attend to these features and emit halfplane parameters $\{\mathbf{n}_{k,h}, d_{k,h}, \mathbf{t}_k, \beta_k\}$ for $K$ convex polytopes $\mathcal{C}_k$; their union is the predicted occupancy, from which per-convex meshes are extracted for simulator use. During training (top), the latent comes from the frozen encoder $\mathcal{E}$ instead.}
  \label{fig:pipeline}
  \vspace{-6pt}
\end{figure}

\ours maps a single RGB image $\mathbf{I}$ of an object to $K$ convex polytopes $\{\mathcal{C}_k\}_{k=1}^{K}$ whose union approximates the object. The core idea is to leave image-to-shape reasoning to a pretrained generator and learn only the readout (Fig.~\ref{fig:pipeline}):
\begin{equation}
  \{\mathcal{C}_k\}_{k=1}^{K} \;=\; \underbrace{g_\theta}_{\substack{\text{trained}\\\text{head}}}\!\big(\underbrace{\mathcal{D}(z)}_{\substack{\text{frozen}\\\text{decoder}}}\big),
  \qquad z \sim \underbrace{p(z \mid \mathbf{I})}_{\substack{\text{frozen}\\\text{DiT}}}.
  \label{eq:overview}
\end{equation}
A frozen diffusion transformer (DiT) samples a shape latent $z$ from the image, a frozen decoder $\mathcal{D}$ expands it into shape features, and a small trained head $g_\theta$ turns them into convexes.

\subsection{Convex Shape Representation}
\label{sec:convex_rep}
Following CvxNet~\cite{deng2020cvxnet}, a shape is a union of $K$ convex polytopes, each an intersection of $H$ halfplanes. Convex $k$ is defined by a translation $\mathbf{t}_k\in\mathbb{R}^3$, unit normals $\mathbf{n}_{k,h}\in\mathbb{S}^2$, and offsets $d_{k,h}\in[-0.5,0]$. With the signed distance $s_{k,h}(\mathbf{x})=\mathbf{n}_{k,h}^{\top}(\mathbf{x}-\mathbf{t}_k)+d_{k,h}$ to halfplane $h$, a point $\mathbf{x}$ lies in $\mathcal{C}_k$ iff $s_{k,h}(\mathbf{x})\le 0$ for all $h$.

\textbf{Differentiable occupancy.}
This membership test is not differentiable, so for learning we relax it with a LogSumExp soft intersection, a sigmoid indicator, and a hard-max union,
\begin{align}
  \hat{s}_k(\mathbf{x}) &= -\tfrac{1}{\beta_k}\log\textstyle\sum_{h=1}^{H}\exp\!\big(\beta_k\, s_{k,h}(\mathbf{x})\big), \\
  o_k(\mathbf{x}) &= \sigma\big(\alpha\,\hat{s}_k(\mathbf{x})\big),\qquad
  O(\mathbf{x}) = \max_{k} o_k(\mathbf{x}),
  \label{eq:union}
\end{align}
where $\sigma$ is the logistic sigmoid, $\alpha=75$ is a fixed sharpness, and $\beta_k\in[50,300]$ is a predicted per-convex blending sharpness. The relaxation is tight and one-sided: LogSumExp bounds give $\hat{s}_k\le-\max_h s_{k,h}\le\hat{s}_k+\log H/\beta_k$, so $o_k\to\mathbf{1}[\mathbf{x}\in\mathcal{C}_k]$ pointwise as $\alpha,\beta_k\to\infty$ and $o_k$ only ever \emph{under}-estimates membership. At $\alpha{=}75$ and $H{=}12$, a point $0.1$ outside a part receives $o_k\le\sigma(-7.5)\approx5{\times}10^{-4}$, while a point $0.1$ inside receives $o_k\gtrsim0.98$ at the smallest $\beta_k{=}50$ (${\ge}0.999$ at $\beta_k{=}300$). Our default uses $K=H=12$.

\textbf{Guarantees.}
Each $\mathcal{C}_k$ is an intersection of closed half-spaces and hence convex, the property that GJK/EPA-based engines require. It is non-empty, since $s_{k,h}(\mathbf{t}_k)=d_{k,h}\le 0$ places $\mathbf{t}_k$ inside. Boundedness is \emph{not} implied by the parameterization: $\mathcal{C}_k$ is bounded iff its recession cone is trivial, \ie iff the normals $\{\mathbf{n}_{k,h}\}_h$ \emph{positively} span $\mathbb{R}^3$. We therefore mesh every convex inside the cube $[-0.6,0.6]^3$ (\S\ref{sec:inference}), which is equivalent to adding the six cube halfplanes and bounds each exported part by construction. Empirically the clipping is never active: an exact recession-cone test over all $227{\times}12=2{,}724$ parts predicted on our benchmark (before the export-time dropping of \S\ref{sec:inference}) finds zero unbounded parts; the origin lies strictly inside the convex hull of each part's normals, with margin $\ge 0.12$ (median $0.53$).

\subsection{Pretrained Image-to-Shape Backbone}
\label{sec:backbone}
We use Hunyuan3D-2-mini~\cite{zhao2025hunyuan3d} as a fixed backbone with two frozen components. The \textbf{image-conditioned DiT} is a flow-matching transformer that samples a shape latent $z\in\mathbb{R}^{L\times c}$ ($L{=}512$ tokens, $c{=}64$) from the image; we use the released weights without fine-tuning. The \textbf{ShapeVAE decoder} $\mathcal{D}$, a linear projection followed by a 16-layer self-attention transformer of width $d{=}1024$, maps $z$ to per-token shape features $f\in\mathbb{R}^{L\times d}$. We discard Hunyuan's SDF readout and attach our head in its place. Because both components are frozen, training never updates a billion-parameter model and never runs the DiT: we encode each training mesh once with the frozen Hunyuan encoder $\mathcal{E}$ (canonical $5120{+}5120$ random/sharp-edge surface samples) and cache the latents.

\subsection{Convex Head}
\label{sec:head}
The head $g_\theta$ is the only trained component. It predicts all $K$ parameter tuples from $f$ in one pass, using one learned token per convex. These $K$ \emph{convex-slot} tokens have width $d$ and are initialized from $\mathcal{N}(0,0.02^2)$. Two residual cross-attention blocks let the slots query $f$, reusing Hunyuan's attention layout and QK normalization, and one self-attention layer over the slots coordinates part assignments. A per-slot LayerNorm and linear readout $\mathbb{R}^d\to\mathbb{R}^{3+2H+H+1}$ produces the translation $\mathbf{t}_k$, the normals via $\tanh$-bounded spherical angles, the offsets $d_{k,h}=-0.5\,\sigma(\cdot)$, and the sharpness $\beta_k=50+250\,\sigma(\cdot)$, so the parameter ranges of \S\ref{sec:convex_rep} hold by construction. The readout is initialized with small weights ($10^{-3}$) and angle biases spread uniformly in $[-0.9,0.9]$, so initial convexes are centered, non-degenerate, and cover the sphere of normals. At $K{=}H{=}12$ the head has ${\sim}38$\,M parameters, a small fraction of the backbone.

\subsection{Training}
\label{sec:training}
We train only the head, on cached latents of artist-created MolmoSpaces meshes~\cite{kim2026molmospaces}. Each mesh is centered and scaled so that its longest bounding-box edge has unit length. Every shape then lies in $[-0.5,0.5]^3$ (the convention of the offsets $d_{k,h}$), with a $0.1$ margin to the export cube of \S\ref{sec:inference}.

\textbf{Occupancy loss.}
For each mesh we precompute $4096$ query points, half near the surface (Gaussian-perturbed surface samples) and half uniformly in $[-0.6,0.6]^3$, with binary labels $O_j^*$ from the watertight mesh, and draw $M{=}2048$ of them per step. Only $p\approx 6\%$ of queries are inside, so we supervise $O_j=O(\mathbf{x}_j)$ with a class-balanced BCE,
\begin{equation}
  \mathcal{L}_{\mathrm{BCE}} = -\tfrac{1}{M}\textstyle\sum_{j}\big[w_+ O_j^*\log O_j + (1-O_j^*)\log(1-O_j)\big],
\end{equation}
with $w_+=(1-p)/p$. This equals $(1-p)$ times the class-mean BCE and thus equalizes the gradient contribution of occupied and free space; without it the model collapses to predicting empty space (\S\ref{sec:exp_ablation}).

\textbf{Convex-pair contrastive loss.}
Occupancy supervises the union but not how it is split into parts, and the head tends toward \emph{slab-stacking}, where many flat parts tile one region. We suppress it with the containment principle of Yang~\etal~\cite{learningconvexdecomp2025}: two points can share a convex only if the segment between them stays inside the shape. For each mesh we precompute $256$ triplets: an anchor $\mathbf{x}_a$ on the surface; a positive $\mathbf{x}_p$ at the first surface re-entry of an inward ray from $\mathbf{x}_a$, so segment $\overline{\mathbf{x}_a\mathbf{x}_p}$ stays inside the volume and the pair \emph{can} share a convex; and $32$ negatives $\mathbf{x}_n^{(i)}$ whose segment to $\mathbf{x}_a$ exits the volume, so they cannot. Each step uses $A{=}64$ anchors with $N{=}16$ negatives each. Unlike Yang~\etal, we need no auxiliary feature field: the loss acts on the head's per-convex responsibilities. With $\mathbf{r}(\mathbf{x})$ the L2-normalized vector $[o_1(\mathbf{x}),\dots,o_K(\mathbf{x})]$ and temperature $\tau{=}0.1$,
\begin{equation}
  \mathcal{L}_{\mathrm{NCE}} = -\tfrac{1}{2A}\textstyle\sum_{a}\Big[\log\tfrac{e^{\mathbf{r}_a^\top\mathbf{r}_p/\tau}}{e^{\mathbf{r}_a^\top\mathbf{r}_p/\tau}+\sum_{i}e^{\mathbf{r}_a^\top\mathbf{r}_n^{(i)}/\tau}} + (a\leftrightarrow p)\Big],
\end{equation}
where $(a\leftrightarrow p)$ swaps anchor and positive.

\textbf{Optimization.}
The objective is $\mathcal{L}=\mathcal{L}_{\mathrm{BCE}}+0.1\,\mathcal{L}_{\mathrm{NCE}}$, optimized with AdamW (lr $3{\times}10^{-4}$, batch $256$, 5 warmup epochs then cosine decay to $10^{-6}$, gradient clipping at $1.0$, EMA $0.999$). $\alpha{=}75$ follows CvxNet and $\tau{=}0.1$ follows Yang~\etal; neither is tuned. The reported head is the validation-best checkpoint at epoch $210$ (${\sim}9$ GPU-hours); ablations use a matched $100$-epoch budget (${\sim}4$ GPU-hours).

\subsection{Inference and Export}
\label{sec:inference}
At test time a single image is processed end-to-end in ${\sim}0.5$\,s on one RTX~5000 Ada GPU: background removal with Hunyuan's bundled remover ($124$\,ms, CPU), DiT sampling with $10$ flow-matching steps and guidance $5.0$ ($409$\,ms), the frozen ShapeVAE decoder ($6$\,ms), and the convex head ($0.4$\,ms). For simulator use, each convex is meshed independently by marching cubes on $o_k$ at $32^3$ resolution within $[-0.6,0.6]^3$ (iso-level $0.1$; $4.4$\,ms per object). Parts whose maximum occupancy is below $0.5$ are dropped, which removes $13$ of the $2{,}724$ parts predicted on our benchmark ($11.94$ parts per object on average). Because every exported part is convex (\S\ref{sec:convex_rep}), engines such as MuJoCo~\cite{todorov2012mujoco}, PyBullet~\cite{coumans2016pybullet}, Genesis~\cite{Genesis}, and Isaac~Lab~\cite{mittal2025isaaclab} accept it as-is: no repair, decimation, or decomposition is required.

\section{Experiments}
\label{sec:experiments}

We evaluate geometric fidelity (\S\ref{sec:exp_geometric}), simulation-readiness across four engines (\S\ref{sec:exp_simready}), pick-and-place on a physical xArm7 (\S\ref{sec:exp_planning}), and ablations (\S\ref{sec:exp_ablation}). Runtimes are measured on an AMD~7960X workstation with $256$\,GB RAM and one RTX~5000 Ada GPU.

\subsection{Geometric Fidelity}
\label{sec:exp_geometric}

\textbf{Data.}
The head is trained on $121{,}056$ artist-created MolmoSpaces meshes~\cite{kim2026molmospaces} (\S\ref{sec:training}). We test single-image generalization on $n{=}227$ held-out objects: $127$ real scans from OmniObject3D~\cite{wu2023omniobject3d} and $100$ watertight household objects from Google Scanned Objects (GSO)~\cite{downs2022googlescannedobjectshighquality}. The dataset's $0$-th rendered view is the input and the original mesh the ground truth. Both sets are disjoint from MolmoSpaces.

\textbf{Baselines.}
No open-world method predicts convex decompositions from an image (\S\ref{sec:related}), so we compare to \emph{reconstruct-then-decompose} pipelines that pair two image-to-3D models with four decomposers. The generators are \textbf{HY3D} (Hunyuan3D-2-mini~\cite{zhao2025hunyuan3d}, $10$ flow-matching steps, the same DiT we adapt) and \textbf{SAM3D}~\cite{chen2025sam}. The classical decomposers are \textbf{CoACD}~\cite{wei2022approximate}, with concavity $0.05$ and at most $12$ hulls, and \textbf{V-HACD}~\cite{mamou2009simple}, with at most $12$ hulls and resolution $10^5$. All other parameters are library defaults (no decimation or vertex cap), and no mesh repair is applied unless stated. The learned decomposers are not mesh-cutting algorithms, so we specify their interface. \textbf{CvxNet}~\cite{deng2020cvxnet} consumes a $4{,}096$-point surface cloud sampled from the reconstructed mesh (normalized to $[-0.5,0.5]^3$) and is retrained on our MolmoSpaces training set, since no open-world weights exist. \textbf{BSP-Net}~\cite{chen2020bsp} consumes a $64^3$ occupancy voxelization and runs the official pretrained checkpoint with the official redundant-convex filtering. Runtimes include these conversions.

\textbf{Generated-mesh quality.}
The meshes these pipelines must decompose are far from clean. Of the $127$ HY3D meshes (median $1.6$\,M faces), $98\%$ contain non-manifold edges (median $7$ per mesh), only $2\%$ pass a watertightness test, $6\%$ have open boundaries, and $59\%$ consist of several components (median $3$, the extra ones being fragments with ${<}0.1\%$ of the vertices). These defects are small in extent, but they disqualify the mesh for tools requiring manifold input.

\textbf{Metrics.}
Each prediction is aligned to its ground truth by centering, unit-diagonal scaling, PCA alignment over the four sign ambiguities, and $50$ iterations of scale-enabled ICP~\cite{besl1992method}. We report \textbf{volumetric IoU} from $10^5$ uniform samples (containment at pitch $\mathrm{extent}/64$). The reference is the mesh itself for watertight GSO objects and its convex hull for open OmniObject3D scans; predictions always use their actual geometry. We also report symmetric L1 \textbf{Chamfer distance} (CD) and \textbf{F-score} at $\tau{=}1\%$ of the ground-truth extent on $10^4$ surface samples, and end-to-end \textbf{runtime}.

\begin{table}[t]
\centering
\caption{\textbf{Single-image convex decomposition} on OmniObject3D\,$+$\,GSO ($n{=}227$), mean\,$\pm$\,std over objects, with IoU also split by dataset (OmniObject3D uses a hull-proxy reference, GSO the exact mesh). $t_\mathrm{total}$ for baselines is mesh generation ($1.56$\,s HY3D, $13.30$\,s SAM3D) plus decomposition. $^{\ast}$GSO means for baselines are recovered from the pooled and OmniObject3D means (exact up to rounding).}
\label{tab:main_results}
\scriptsize
\setlength{\tabcolsep}{3pt}
\resizebox{\linewidth}{!}{%
\begin{tabular}{lcccccc}
\toprule
Method & IoU (\%)\,$\uparrow$ & \multicolumn{1}{c}{Omni} & \multicolumn{1}{c}{GSO} & F (\%)\,$\uparrow$ & CD (mm)\,$\downarrow$ & $t_\mathrm{total}$ (s)\,$\downarrow$ \\
\midrule
HY3D + CoACD  & $62.8\pm19.0$ & $62.9$ & $62.7^{\ast}$ & $16.4\pm12.9$ & $6.67\pm7.07$ & $7.82$ \\
HY3D + V-HACD & $61.0\pm18.6$ & $61.1$ & $60.8^{\ast}$ & $16.1\pm15.3$ & $6.76\pm6.77$ & $7.50$ \\
HY3D + CvxNet & $58.7\pm15.9$ & $56.1$ & $62.0^{\ast}$ & $13.2\pm10.9$ & $8.17\pm11.63$ & $3.05$ \\
HY3D + BSP-Net & $59.7\pm16.4$ & $59.7$ & $59.7^{\ast}$ & $14.4\pm11.3$ & $8.73\pm9.46$ & $3.67$ \\
SAM3D + CoACD & $61.5\pm20.2$ & $61.3$ & $62.0$ & $\mathbf{19.2\pm14.7}$ & $\mathbf{5.87\pm6.82}$ & $19.04$ \\
SAM3D + V-HACD & $62.2\pm18.3$ & $61.0$ & $63.8^{\ast}$ & $17.1\pm14.8$ & $6.12\pm6.78$ & $13.76$ \\
SAM3D + CvxNet & $60.0\pm15.0$ & $57.3$ & $63.4^{\ast}$ & $13.5\pm10.8$ & $7.28\pm7.53$ & $13.35$ \\
SAM3D + BSP-Net & $61.8\pm14.4$ & $62.1$ & $61.5^{\ast}$ & $14.9\pm11.2$ & $8.41\pm11.87$ & $14.09$ \\
\midrule
\textbf{\ours (ours)} & $\mathbf{65.7\pm13.2}$ & $\mathbf{63.1}$ & $\mathbf{68.9}$ & $15.6\pm10.6$ & $6.28\pm6.31$ & $\mathbf{0.51\pm0.04}$ \\
\bottomrule
\end{tabular}%
}
\vspace{-10pt}
\end{table}

\begin{figure*}[t]
\centering
\includegraphics[width=0.92\textwidth]{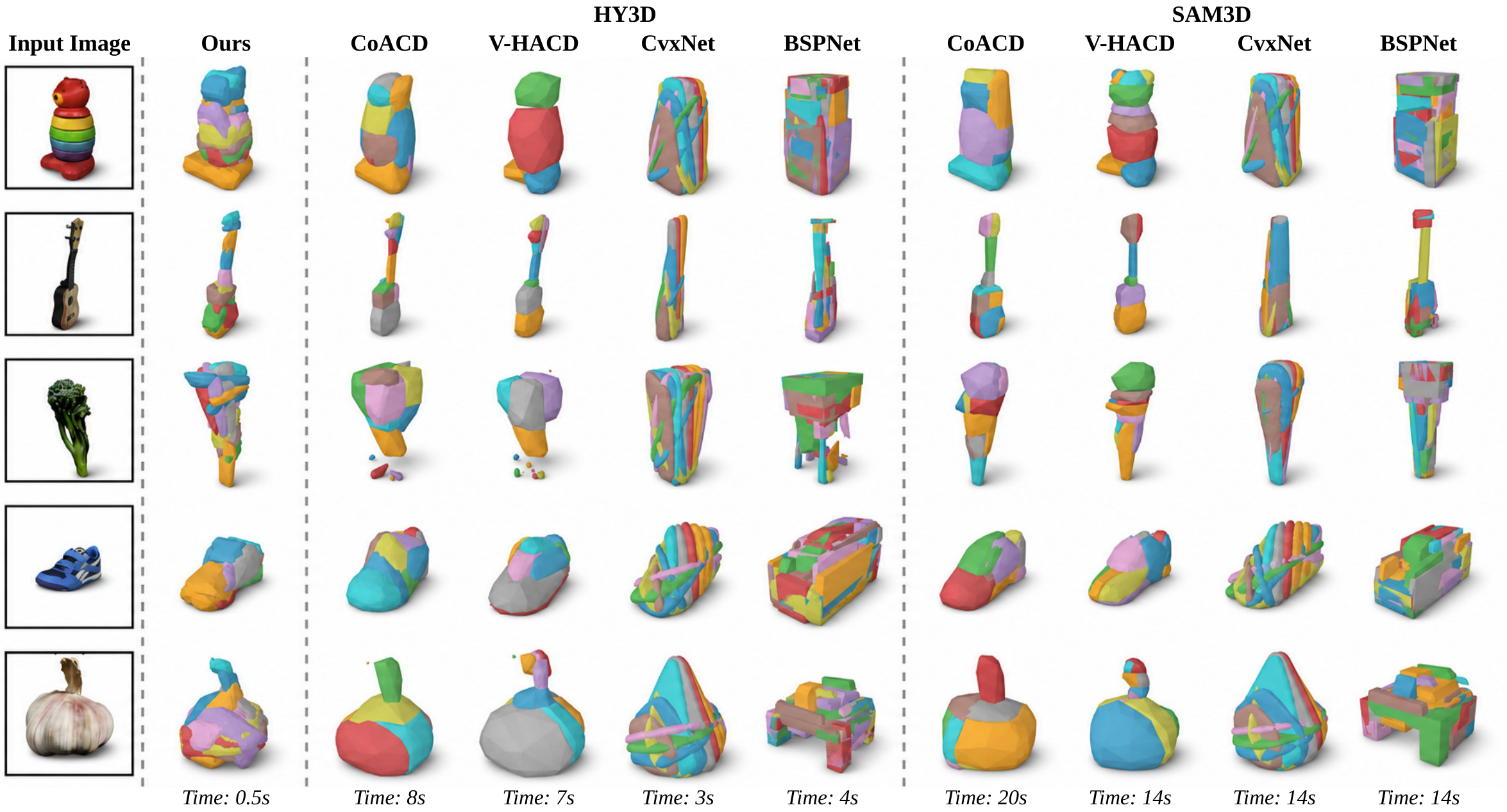}
\vspace{-6pt}
\caption{\textbf{Qualitative comparison.} Each color is one convex part; the time under each column is the rounded mean end-to-end runtime. \ours produces compact, volumetrically faithful decompositions in one pass, whereas reconstruct-then-decompose pipelines inherit and amplify reconstruction artifacts into fragmented or misaligned parts.}
\label{fig:qualitative}
\vspace{-3pt}
\end{figure*}

\textbf{Results.}
Table~\ref{tab:main_results} and Fig.~\ref{fig:qualitative} summarize the comparison. \ours attains the highest pooled IoU ($65.7\%$), about $3$ points above the strongest pipeline (HY3D+CoACD, $62.8\%$), with the lowest variance of any method. It does so in $0.51$\,s end-to-end: $3\times$ faster than HY3D mesh generation \emph{alone} and $6$--$37\times$ faster than the complete pipelines ($3.05$--$19.04$\,s).

\textbf{Per-dataset analysis.}
On the $100$ watertight GSO objects, where IoU is exact, the margin widens to $5$--$9$ points over every pipeline ($68.9$ vs.\ ${\le}63.8\%$); against SAM3D+CoACD, for which we hold per-object GSO results, a paired Wilcoxon test gives $p{=}0.017$. On the $127$ OmniObject3D scans, whose open surfaces force a convex-hull reference, \ours and the best pipeline are tied ($63.1$ vs.\ $62.9\%$, $p{=}0.10$). A hull reference rewards inflation and hides concavity errors, so this is the split that flatters the pipelines. The advantage of \ours is robustness rather than per-object dominance: the CoACD/V-HACD pipelines beat \ours on $55$--$65\%$ of OmniObject3D objects by a median of $3$--$4$ points, but collapse to IoU $<40\%$ on $14$--$17\%$ of objects versus $5\%$ for \ours.

\textbf{Sampling noise.}
Over four DiT sampling seeds, \ours scores $62.1\pm0.7$ IoU on OmniObject3D ($63.1$, $61.5$, $62.2$, $61.7$; Table~\ref{tab:main_results} reports the first), with a median per-object spread of $1.7$ points. This sets the noise floor for the ablations below.

\textbf{Surface metrics and mesh repair.}
Surface metrics favor methods that preserve local detail: SAM3D+CoACD has the best F-score and CD, at the highest IoU variance ($20.2\%$), and \ours trails its F-score by $3.6$ points (\S\ref{sec:conclusion}). Mesh repair does not close the IoU gap to \ours. SAM3D meshes add double layers and non-watertight components to the defects listed above. Repairing and simplifying them before CoACD (duplicate/non-manifold removal, hole filling, quadric decimation) raises IoU from $61.5$ to $63.2\%$, still below \ours, while decomposition time more than doubles ($6.1\to13.8$\,s per object) because manifold inputs route CoACD to its slower exact path. Pre-processing narrows the quality gap only by widening the runtime gap.

\subsection{Simulation-Readiness Across Physics Engines}
\label{sec:exp_simready}

``Simulation-ready'' is usually asserted by showing that an asset loads. This is not enough: engines accept almost any mesh and, for dynamic bodies, silently replace non-convex geometry with a proxy. We therefore load the $127$ OmniObject3D objects into \textbf{MuJoCo}~\cite{todorov2012mujoco}, \textbf{PyBullet}~\cite{coumans2016pybullet}, \textbf{Genesis}~\cite{Genesis}, and \textbf{Isaac Sim} (PhysX)~\cite{mittal2025isaaclab}. \textbf{Ours} is loaded as one convex hull per predicted part, the engines' hull mode for mesh colliders. The \textbf{raw} HY3D mesh (${\sim}2$\,M faces) is loaded under each engine's \textbf{default} mesh handling and, where offered, under its non-convex alternatives: PyBullet's concave triangle mesh, PhysX's signed-distance-field (SDF) collider, and PhysX's built-in convex decomposition (V-HACD). Each object is normalized to $0.2$\,m, dropped from $0.1$\,m onto the ground, and simulated for $10$\,s at $2$\,ms steps. Genesis runs on CPU; its load times include per-scene kernel compilation.

\textbf{Metrics.}
\emph{Asset preserved} is the fraction of objects whose engine-effective collision geometry has voxel IoU $\ge 0.9$ with the asset as handed to the engine. The engine-effective geometry is the geoms and cooked hulls read back from Genesis and PhysX; the exact convex hull(s) of the loaded mesh(es) for MuJoCo, PyBullet, and PhysX's default; and the exact mesh for SDF/concave modes. This is a round-trip measure of what the engine does to an asset, not of fidelity to the object (Table~\ref{tab:main_results}). We also report median \emph{load time}, from mesh files to a stepped model with cold caches; median engine-reported contact \emph{penetration} at rest; and the fraction \emph{at rest}, \ie whose pose changes by less than $1$\,mm and $1^{\circ}$ over the final second.

\begin{table}[t]
\centering
\caption{\textbf{Simulation-readiness beyond loading} ($127$ OmniObject3D objects, four engines). ``Raw'' is the HY3D mesh; ``default'' is the engine's untouched mesh handling. ``Preserved'': round-trip IoU$\ge0.9$ between engine-effective and input geometry (not fidelity to the object). All rows load $100\%$ of objects except $^{\ddagger}$, where PhysX cooking crashed on $1$ (hull) and $6$ (SDF) objects, excluded from that row. Genesis's SDF collider is omitted: in a $20$-object pilot it exceeded $180$\,GB RAM or $15$\,min on $13$ objects and took $1$--$6$\,min on the rest. $^{\dagger}$exact shape, but dynamic concave meshes register zero object--object contacts in PyBullet.}
\label{tab:simready}
\scriptsize
\setlength{\tabcolsep}{2.5pt}
\begin{tabular}{llcccc}
\toprule
Engine & Geometry & Preserved & Load & Penetr. & At rest \\
 & & (\%) & (s) & (mm) & (\%) \\
\midrule
\multirow{2}{*}{MuJoCo} & \textbf{ours} & \SRmjOursC & \SRmjOursL & \SRmjOursP & \SRmjOursR \\
 & raw, default ($\to$\,hull) & \SRmjRawC & \SRmjRawL & \SRmjRawP & \SRmjRawR \\
\midrule
\multirow{3}{*}{PyBullet} & \textbf{ours} & \SRpbOursC & \SRpbOursL & \SRpbOursP & \SRpbOursR \\
 & raw, default ($\to$\,hull) & \SRpbRawC & \SRpbRawL & \SRpbRawP & \SRpbRawR \\
 & raw, concave mesh & 100$^{\dagger}$ & \SRpbConL & \SRpbConP & \SRpbConR \\
\midrule
\multirow{2}{*}{Genesis} & \textbf{ours} & \SRgeOursC & \SRgeOursL & \SRgeOursP & \SRgeOursR \\
 & raw, default (decim.$+$CoACD) & \SRgeRawC & \SRgeRawL & \SRgeRawP & \SRgeRawR \\
\midrule
\multirow{4}{*}{Isaac Sim} & \textbf{ours} & \SRisOursC & \SRisOursL & \SRisOursP & \SRisOursR \\
 & raw, default ($\to$\,hull)$^{\ddagger}$ & \SRisRawC & \SRisRawL & \SRisRawP & \SRisRawR \\
 & raw, V-HACD decomp. & \SRisCdC & \SRisCdL & \SRisCdP & \SRisCdR \\
 & raw, SDF$^{\ddagger}$ & 100 & \SRisSdfL & \SRisSdfP & \SRisSdfR \\
\bottomrule
\end{tabular}
\vspace{-10pt}
\end{table}

\textbf{Results.}
Table~\ref{tab:simready} exposes what ``it loads'' hides. \emph{(1)~Default handling changes the shape}: MuJoCo, PyBullet, and PhysX all replace the dynamic raw mesh by its single convex hull (PhysX logs a warning; the others are silent). This is the wrong collision shape for about $\SRrawWrong\%$ of objects, and it is exactly the auto-hull geometry that fails $95$ of $100$ trials in \S\ref{sec:exp_planning}. The single hull is also a poor \emph{dynamic} proxy: it keeps rocking on $16$--$21\%$ of objects in MuJoCo and PyBullet. \emph{(2)~Built-in decomposition does not recover it}: Genesis decimates to $500$ faces and runs CoACD when the hull error exceeds $15\%$ (\SRgeDecomp\ of $127$ objects), \ie it re-introduces the very pipeline we bypass, yet preserves the asset for only $\SRgeRawC\%$ of objects while paying seconds of preprocessing per object. PhysX's built-in V-HACD fares no better (\SRisCdC\% preserved): approximate decomposition of a $2$\,M-face non-manifold mesh under engine defaults is not a substitute for clean input. \emph{(3)~Non-convex colliders are exact but not free}: PyBullet's concave mode registers \emph{zero} object--object contacts between dynamic bodies, so clutter cannot interact. PhysX's SDF collider costs $\SRisSdfL$\,s of cooking per object, crashes on $6$ of them, and leaves $14\%$ of objects (mostly round fruit and balls) slowly rolling. Genesis's SDF collider is prohibitive on these meshes (Table~\ref{tab:simready}, caption). \emph{(4)~\ours geometry is used as delivered}: it needs no preprocessing in any engine and is preserved on $\ge\SRoursMinC\%$ of objects everywhere (one consistent miss: a thin wall clock). It loads in tens to hundreds of milliseconds in MuJoCo, PyBullet, and PhysX, settles at sub-millimeter penetration, and comes to rest in $\ge\SRoursMinR\%$ of drops (exception: a coiled belt creeping in MuJoCo).

\textbf{Complete simulation assets.}
When a visual mesh is also needed, it and the convexes decode from the same frozen DiT latent and the head adds $0.4$\,ms, so a complete asset costs ${\sim}1.6$\,s versus $7.8$\,s for HY3D+CoACD, at higher collision IoU ($65.7$ vs.\ $62.8\%$). If only collision geometry is needed, as on our robot, the full $6$--$37\times$ speedup applies.

\subsection{Pick-and-Place on a Physical xArm7 in Clutter}
\label{sec:exp_planning}

We next test \ours where it must ultimately work: planning and executing through clutter on a real robot. Per-object metrics miss that under-covered geometry permits unsafe plans, that over-thickened geometry (\eg an inflated hull) blocks grasps and approach paths, and that every stage from capture to a planner-ready scene consumes planning budget.

\begin{figure*}[t]
\centering
\includegraphics[width=0.95\textwidth]{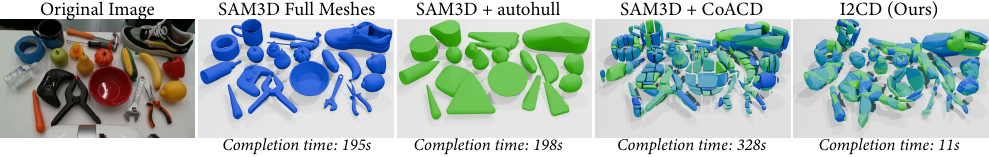}
\vspace{-3pt}
\caption{\textbf{Real cluttered scene.} From one RGB-D observation (left), SAM3D reconstructs dense per-object meshes; auto-hull collapses each to one convex hull; SAM3D+CoACD decomposes post hoc; \ours predicts compact convex geometry directly. Times are wall-clock to a planner-ready $20$-object scene.}
\label{fig:comparison}
\vspace{-3pt}
\end{figure*}

\begin{figure}[t]
\centering
\includegraphics[width=0.92\linewidth]{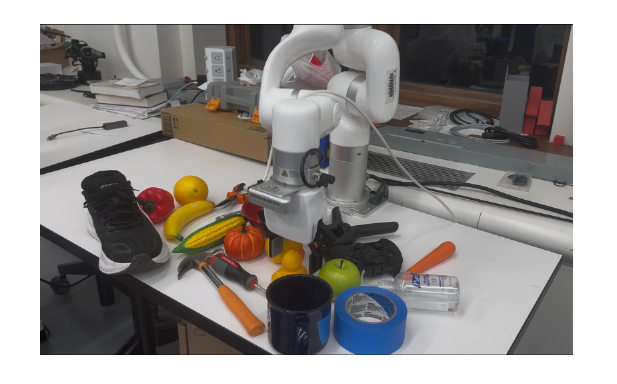}
\vspace{-3pt}
\caption{\textbf{Physical execution.} The xArm7 executing a planned pick in the $20$-object scene of Fig.~\ref{fig:comparison}. Every trial in Table~\ref{tab:planning} is run on this robot.}
\label{fig:realworld}
\vspace{-1pt}
\end{figure}

\textbf{Setup and protocol.}
A calibrated RealSense on a 7-DoF UFactory xArm7 (Fig.~\ref{fig:realworld}) captures a single RGB-D frame of a tabletop with $20$ household and tool objects. SAM3~\cite{carion2025sam3segmentconcepts} segments the image, each method predicts object-centric convex geometry from the masked crops, and the geometries are placed at their depth-registered poses in a MuJoCo scene with the same arm and a table plane (Fig.~\ref{fig:comparison}). Each of the $20$ objects serves in turn as the pick target with the other $19$ as obstacles. RRT-Connect~\cite{kuffner2000rrt} plans a four-phase \textsc{approach}$\to$\textsc{grasp}$\to$\textsc{retreat}$\to$\textsc{place} motion in the 7-DoF configuration space with a $5$\,s timeout per phase, and \emph{the planned trajectory is executed on the physical robot}. A trial succeeds only if all four phases are planned within budget and the execution completes the pick-and-place without collision; any IK, grasp-pose, planning, or execution failure counts as failure. Each target is repeated with $5$ planner seeds, for $100$ physical trials per method. We report the time $t_\mathrm{geom}$ to planner-ready geometry for all $20$ objects, the one-time MuJoCo load time $t_\mathrm{load}$, and execution success; per-query collision-check time differs by ${\le}0.08$\,ms across methods.

\textbf{Baselines.}
Segmentation, pose registration, scene assembly, planner, and robot are held fixed; only the geometry changes. \textbf{SAM3D + auto-hull} replaces each SAM3D mesh with a single convex hull, the engine-default proxy of \S\ref{sec:exp_simready}. \textbf{SAM3D + CoACD} is the pipeline with the best surface metrics in Table~\ref{tab:main_results}, with the same CoACD settings. We chose it because grasp and approach feasibility depend on local surface accuracy. It is also the slowest: the IoU-best pipeline, HY3D+CoACD, was not run on the robot, but at $7.8$\,s per object it would need about $156$\,s for the scene, so the speedup would be ${\sim}14\times$ against it rather than ${\sim}30\times$. \textbf{\ours} predicts $K{=}12$ convexes per object with no decomposition stage.

\begin{table}[t]
\centering
\caption{\textbf{Physical pick-and-place in a $20$-object scene.} $t_\mathrm{geom}$: time to planner-ready geometry for all objects; $t_\mathrm{load}$: one-time MuJoCo load. Success over $20$ targets\,$\times$\,$5$ seeds, executed on the xArm7.}
\label{tab:planning}
\small
\setlength{\tabcolsep}{5pt}
\begin{tabular}{lccc}
\toprule
Method & $t_\mathrm{geom}$ (s)\,$\downarrow$ & $t_\mathrm{load}$ (s)\,$\downarrow$ & Success\,$\uparrow$ \\
\midrule
SAM3D + auto-hull & $198$ & $0.9$ & $\;\;5/100$ \\
SAM3D + CoACD     & $328$ & $1.7$ & $\mathbf{90/100}$ \\
\midrule
\textbf{\ours (ours)} & $\mathbf{11}$ & $\mathbf{0.9}$ & $85/100$ \\
\bottomrule
\end{tabular}
\vspace{-1pt}
\end{table}

\textbf{Results.}
Table~\ref{tab:planning} reports the results. \emph{(1)~Geometry quality gates task success}: the single-hull baseline succeeds in only $5/100$ trials, since inflated hulls block valid grasps and approach paths in clutter even though each hull is a faithful outer bound of its object. Multi-convex geometry succeeds in $90/100$ (SAM3D+CoACD) and $85/100$ (\ours). The five-trial difference is not significant even if the $100$ trials were independent (Fisher's exact test, $p{=}0.39$), and less so given that they are $5$ seeds on each of only $20$ configurations. The remaining failures trace mostly to pose-registration error shifting tightly packed obstacles into the grasp envelope, plus, for \ours, slightly lower local surface fidelity from direct $K{=}12$ prediction. \emph{(2)~The pipeline, not the planner, is the bottleneck}: \ours delivers planner-ready geometry for the scene in $11$\,s versus $328$\,s for SAM3D+CoACD, a ${\sim}30\times$ reduction (${\sim}14\times$ against HY3D+CoACD, see above), while load and collision-check times are tied.

\subsection{Ablations}
\label{sec:exp_ablation}

\begin{table}[t]
\centering
\caption{\textbf{Training objective} on OmniObject3D ($n{=}127$); each row is that objective's validation-best checkpoint (ours is the Table~\ref{tab:main_results} model; its $63.8$ vs.\ $63.1$ there is DiT sampling noise).}
\label{tab:loss_ablation}
\footnotesize
\setlength{\tabcolsep}{5pt}
\begin{tabular}{lccc}
\toprule
Loss & IoU (\%)\,$\uparrow$ & F (\%)\,$\uparrow$ & CD (mm)\,$\downarrow$ \\
\midrule
BCE\,$+$\,NCE (ours) & $\mathbf{63.8}$ & $\mathbf{12.5}$ & $\mathbf{5.98}$ \\
Class-balanced BCE only & $62.2$ & $11.4$ & $6.49$ \\
Unbalanced BCE & $54.7$ & $9.7$ & $8.89$ \\
MSE & \multicolumn{3}{c}{\emph{failed:} $127/127$ empty} \\
\bottomrule
\end{tabular}
\vspace{-5pt}
\end{table}

\begin{table}[t]
\centering
\caption{\textbf{Capacity} on OmniObject3D at a matched $100$-epoch budget.}
\label{tab:capacity}
\scriptsize
\setlength{\tabcolsep}{2.5pt}
\resizebox{\linewidth}{!}{%
\begin{tabular}{@{}cccc@{\hspace{14pt}}lc@{}}
\multicolumn{4}{@{}c@{\hspace{14pt}}}{\textbf{(a) Parts $K$ vs.\ halfplanes $H$}} & \multicolumn{2}{c@{}}{\textbf{(b) Head depth}} \\
\multicolumn{4}{@{}c@{\hspace{14pt}}}{Cell: IoU\,/\,F\,/\,CD} & \multicolumn{2}{c@{}}{(cross-attn, self-attn)} \\
\cmidrule(r{7pt}){1-4}\cmidrule(l{7pt}){5-6}
$K\backslash H$ & $6$ & $12$ & $24$ & Head (layers) & IoU \\
\cmidrule(r{7pt}){1-4}\cmidrule(l{7pt}){5-6}
$6$  & 61.4\,/\,11.9\,/\,6.66 & 65.5\,/\,13.1\,/\,6.52 & 65.3\,/\,12.8\,/\,6.76 & \multirow{1.5}{*}{(1, 0) minimal} & \multirow{1.5}{*}{61.4} \\
$12$ & 61.3\,/\,12.4\,/\,6.79 & 63.1\,/\,13.1\,/\,6.23 & 64.4\,/\,12.8\,/\,6.00 & & \\
$24$ & 58.6\,/\,12.4\,/\,6.49 & 60.2\,/\,12.0\,/\,6.59 & 61.6\,/\,12.6\,/\,6.11 & \multirow{-1.5}{*}{(2, 1) default} & \multirow{-1.5}{*}{62.0} \\
\cmidrule(r{7pt}){1-4}\cmidrule(l{7pt}){5-6}
\end{tabular}}
\vspace{-10pt}
\end{table}

\begin{figure}[t]
\centering
\includegraphics[width=0.95\linewidth]{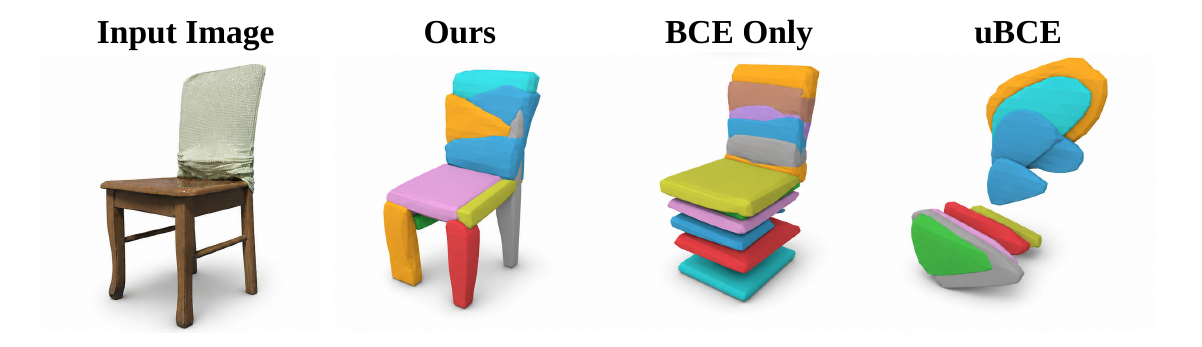}
\vspace{-1pt}
\caption{\textbf{Qualitative ablation of the objective.} \ours yields compact, coherent parts; class-balanced BCE alone tiles the interior with thin slabs; unbalanced BCE (uBCE) under-fills the shape.}
\label{fig:ablation}
\vspace{-3pt}
\end{figure}

\textbf{Objective} (Table~\ref{tab:loss_ablation}, Fig.~\ref{fig:ablation}). We hold architecture and inference fixed and vary only the loss. Class balancing is essential: removing it drops IoU by $7.5$ points, as the ${\sim}94\%$ free-space queries dominate and the model drifts toward empty, under-filled predictions. MSE fails outright because its gradient vanishes in the saturated sigmoid tails. The convex-pair contrastive term improves all three metrics ($+1.6$ IoU, $+1.1$ F, $-0.5$\,mm CD) and, qualitatively, replaces stacks of thin slabs with spatially coherent parts.

\textbf{Capacity} (Table~\ref{tab:capacity}). Sweeping $K,H\in\{6,12,24\}^2$ (nine retrained heads) spans $58.6$--$65.5$ IoU. Fewer, richer parts help ($K{=}6$, $H{=}12$ is best), while $K{=}24$ hurts because slots split thin objects into slivers. We nevertheless keep $K{=}H{=}12$ as the default so that \ours receives the same ${\le}12$-hull budget we give CoACD and V-HACD; $K{=}6$ would widen the margin in Table~\ref{tab:main_results} but make the comparison unequal. A minimal head with one cross-attention and no self-attention layer is within $0.6$ IoU of the default. The two $K{=}H{=}12$ entries ($63.1$ and $62.0$) are the same configuration evaluated in two independent $100$-epoch runs. Their gap matches the $0.7$-point seed spread measured above, so differences below ${\sim}1$ point in this table should be read as noise. Within that resolution, head depth is not the driver of performance; the frozen backbone's prior is.

\textbf{Marching-cubes resolution.} Re-meshing the same predicted parameters at $R\in\{32,64,128\}$ changes IoU by less than $0.15$ points and improves CD by only $0.45$\,mm, while meshing time grows from $4.4$ to $222$\,ms per object. The residual gap to SAM3D+CoACD on surface metrics is thus a property of the predicted geometry, not the export.

\textbf{Train--test latent gap.} The head is trained on latents encoded from ground-truth meshes but receives image-conditioned DiT samples at test time. Feeding the trained head GT-mesh-encoded latents of the $227$ benchmark objects yields $68.9\%$ IoU versus $65.7\%$ with DiT samples. This $3.2$-point gap stems from the backbone's image-conditioned sampling, not the head, and will shrink with stronger backbones.

\section{Conclusion and Limitations}
\label{sec:conclusion}

We presented \ours, which adapts a frozen image-to-3D foundation model to emit convex collision geometry directly from a single RGB image. A 38\,M-parameter head trained in under ten GPU-hours matches or exceeds eight reconstruct-then-decompose pipelines in volumetric IoU at $6$--$37\times$ lower latency. Its geometry is used as delivered by every engine and settles stably in MuJoCo, PyBullet, Genesis, and Isaac Sim without any preprocessing. On a physical xArm7, it cuts the time to a planner-ready cluttered scene by ${\sim}30\times$ at comparable execution success. More broadly, pretrained 3D generators need not be used only through their meshes: their shape features are a substrate for robotics-specific heads that predict what downstream systems consume.

\textbf{Limitations.}
\ours inherits the errors of its frozen backbone (missed thin structures, hallucinated or over-smoothed parts), and the $3.2$-point IoU gap between mesh-encoded and image-sampled latents is entirely a backbone property. Both improve with stronger backbones, and the head is agnostic to which one is used. Surface fidelity trails volumetric fidelity: \ours is $3.6$ F-score points (${\sim}19\%$ relative) behind the best pipeline, and the marching-cubes ablation shows this is a property of the $K{=}12$ predicted geometry, not of export. For quasi-static planning this matters little. In contact-rich tasks such as non-prehensile pushing, where surface detail acts on every contact, it is the limitation most likely to bind; adaptive slot allocation and per-slot existence prediction are the natural remedies. Convex slots group geometry by proximity rather than semantics (they do not separate ``seat'' from ``legs''). Finally, our real-robot study covers one tabletop scene and a quasi-static task. Although its $100$ trials span $20$ target/obstacle configurations, scenes with complex support relations and contact-rich tasks would test the geometry more stringently and are the natural next step.

\bibliographystyle{IEEEtran}
\bibliography{refs}

\end{document}